\documentclass[11pt,a4paper]{article}
\usepackage[hyperref]{acl2019}
\usepackage{times}
\usepackage{latexsym}

\usepackage{url}
\usepackage{color}
\usepackage{algorithm}
\usepackage{algorithmic}
\usepackage{helvet}
\usepackage{courier}
\usepackage{url}
\usepackage{amsmath}
\usepackage{amssymb}
\usepackage{amsopn}
\usepackage{subcaption}
\usepackage{graphicx}
\usepackage{multirow}
\usepackage[english]{babel}
\usepackage{bm}

\usepackage{tikz}
\usepackage{tabularx}
\usepackage{array}
\newcolumntype{C}[1]{>{\centering}p{#1}}

\aclfinalcopy % Uncomment this line for the final submission
\def\aclpaperid{1521} %  Enter the acl Paper ID here

\title{Progressive Self-Supervised Attention Learning for \\ Aspect-Level Sentiment Analysis}

\author{Jialong Tang$^{1,2,3*}$, \ Ziyao Lu$^{1*}$, \  Jinsong Su$^{1\dagger}$, \
Yubin Ge$^{4}$, \ Linfeng Song$^{5}$, \\
\textbf{Le Sun}$^{2}$, \ \textbf{Jiebo Luo}$^{5}$\\
$^{1}$Xiamen University, Xiamen, China\\
$^{2}$Institute of Software, Chinese Academy of Sciences, Beijing, China\\
$^{3}$University of Chinese Academy of Sciences, Beijing, China\\
$^{4}$University of Illinois at Urbana-Champaign, Urbana, IL 61801, USA\\
$^{5}$Department of Computer Science, University of Rochester, Rochester NY 14627, USA\\
 {\tt jialong2019@iscas.ac.cn, ziyaolu2018@stu.xmu.edu.cn} \\
 {\tt jssu@xmu.edu.cn}
}

\date{}

\begin{document}
\maketitle
\begin{abstract}
{
\renewcommand{\thefootnote}{\fnsymbol{footnote}}
\footnotetext[1]{Equal contribution}
\footnotetext[2]{Corresponding author}
}
In aspect-level sentiment classification (ASC),
it is prevalent to equip dominant neural models with attention mechanisms,
for the sake of acquiring the importance of each context word on the given aspect.
However,
such a mechanism tends to excessively focus on a few frequent words with sentiment polarities,
while ignoring infrequent ones.
In this paper,
we propose a progressive self-supervised attention learning approach for neural ASC models,
which automatically mines useful attention supervision information from a training corpus to refine attention mechanisms.
Specifically,
we iteratively conduct sentiment predictions on all training instances.
Particularly,
at each iteration,
the context word with the maximum attention weight is extracted as the one with active/misleading influence on the correct/incorrect prediction of every instance,
and then the word itself is masked for subsequent iterations.
Finally,
we augment the conventional training objective with a regularization term,
which enables ASC models to continue equally focusing on the extracted active context words
while decreasing weights of those misleading ones.
Experimental results on multiple datasets show that our proposed approach yields better attention mechanisms,
leading to substantial improvements over the two state-of-the-art neural ASC models.
%Source code and trained models are available at https://github.com/DeepLearnXMU/PSSAttention.
Source code and trained models are available.\footnote[1]{https://github.com/DeepLearnXMU/PSSAttention} 
\end{abstract}

\section{Introduction}
Aspect-level sentiment classification (ASC),
as an indispensable task in sentiment analysis,
aims at inferring the sentiment polarity of an input sentence in a certain aspect.
In this regard,
previous representative models are mostly discriminative classifiers
based on manual feature engineering,
such as Support Vector Machine \cite{Kiritchenko:SemEval2014,Wagner:SemEval2014}.
Recently,
with the rapid development of deep learning,
dominant ASC models have evolved into neural network (NN) based models \cite{Tang:EMNLP2016,Wang:EMNLP2016,Tang:COLING2016,Ma:IJCAI2017,Chen:EMNLP2017,Li:ACL2018,Wang:ACL2018},
which are able to automatically learn the aspect-related semantic representation of an input sentence
and thus exhibit better performance.
Usually,
these NN-based models are equipped with attention mechanisms to learn the importance of each context word towards a given aspect.
It can not be denied that attention mechanisms play vital roles in
neural ASC models.

However, the existing attention mechanism in ASC suffers from a major drawback.
Specifically,
it is prone to overly focus on a few frequent words with sentiment polarities
and little attention is laid upon low-frequency ones.
As a result,
the performance of attentional neural ASC models is still far from satisfaction.
We speculate that this is because there exist widely ``\emph{apparent patterns}'' and ``\emph{inapparent patterns}''.
Here, ``\emph{apparent patterns}'' are interpreted as high-frequency words with
strong sentiment polarities and ``\emph{inapparent patterns}'' are referred to as low-frequency ones in training data.
As mentioned in \cite{Li:ACL2018,Xu:CONLL2018,Lin:ICCV2017},
NNs are easily affected by these two modes:
``\emph{apparent patterns}'' tend to be overly learned
while ``\emph{inapparent patterns}'' often can not be fully learned.

\newcommand*{\MinNumberb}{0.0}%
\newcommand*{\MaxNumberb}{0.5}%
\newcommand{\Appb}[2]{\pgfmathsetmacro{\PercentColorb}{100.0*(#2-\MinNumberb)/(\MaxNumberb-\MinNumberb)}\colorbox{red!\PercentColorb!white}{\strut #1}}
\begin{table*}[t]
\centering
\small
\begin{tabularx}{13cm}{|p{20pt}<{\centering}|l|X<{\centering}|}
\hline
{\bf Type} & \multicolumn{1}{|c|}{\bf Sentence} & {\bf Ans./Pred.}\\
\hline
\hline
{\bf Train} & \Appb{The}{0.06}\Appb{\bf [place]}{0.08}\Appb{is}{0.06}\Appb{small}{0.27}\Appb{and}{0.12}\Appb{crowded}{0.09}\Appb{but}{0.05}\Appb{the}{0.03}\Appb{service}{0.06}\Appb{is}{0.04}\Appb{quick}{0.13}\Appb{.}{0.01}& {\bf Neg / ---}\\
\hline
{\bf Train} & \Appb{The}{0.06}\Appb{\bf [place]}{0.08}\Appb{is}{0.07}\Appb{a}{0.03}\Appb{bit}{0.07}\Appb{too}{0.15}\Appb{small}{0.35}\Appb{for}{0.08}\Appb{live}{0.05}\Appb{music}{0.05}\Appb{.}{0.01}& {\bf Neg / ---}\\
\hline
{\bf Train} & \Appb{The}{0.02}\Appb{service}{0.03}\Appb{is}{0.03}\Appb{decent}{0.08}\Appb{even}{0.03}\Appb{when}{0.05}\Appb{this}{0.15}\Appb{small}{0.30}\Appb{\bf [place]}{0.15}\Appb{is}{0.05}\Appb{packed}{0.10}\Appb{.}{0.01}& {\bf Neg / ---}\\
\hline
\hline
{\bf Test} & \Appb{At}{0.05}\Appb{lunch}{0.15}\Appb{time}{0.22}\Appb{,}{0.05}\Appb{the}{0.10}\Appb{\bf [place]}{0.18}\Appb{is}{0.10}\Appb{crowded}{0.14}\Appb{.}{0.01}& {\bf Neg / Pos}\\
\hline
{\bf Test} & \Appb{A}{0.05}\Appb{small}{0.28}\Appb{area}{0.12}\Appb{makes}{0.05}\Appb{for}{0.05}\Appb{quiet}{0.10}\Appb{\bf [place]}{0.15}\Appb{to}{0.05}\Appb{study}{0.05}\Appb{alone}{0.1}\Appb{.}{0.01}& {\bf Pos / Neg}\\
\hline
\end{tabularx}
\caption{\label{Table_Example1}
The example of attention visualization for five sentences,
where the first three are training instances and the last two are test ones.
The bracketed bolded words are target aspects.
Ans./Pred. = ground-truth/predicted sentiment label.
Words are highlighted with different degrees according to attention weights.
}
\end{table*}

Here we use sentences in Table \ref{Table_Example1} to explain this defect.
In the first three training sentences,
given the fact that the context word ``\emph{small}" occurs frequently with negative sentiment,
the attention mechanism pays more attention to it and directly relates the sentences containing it with negative sentiment.
This inevitably causes another informative context word ``\emph{crowded}" to be partially neglected in spite of it also possesses negative sentiment.
Consequently,
a neural ASC model incorrectly predicts the sentiment of the last two test sentences:
in the first test sentence,
the neural ASC model fails to capture the negative sentiment implicated by ''\emph{crowded}";
while,
in the second test sentence,
the attention mechanism directly focuses on ``\emph{small}"
though it is not related to the given aspect.
Therefore,
we believe that the attention mechanism for ASC still leaves tremendous room for improvement.

One potential solution to the above-mentioned issue is supervised attention,
which,
however,
is supposed to be manually annotated, requiring labor-intense work.
In this paper,
we propose a novel progressive self-supervised attention learning approach for neural ASC models.
Our method is able to automatically and incrementally mine attention supervision information from a training corpus,
which can be exploited to guide the training of attention mechanisms in ASC models.
The basic idea behind our approach roots in the following fact:
the context word with the maximum attention weight has the greatest impact on the sentiment prediction of an input sentence.
Thus, such a context word of a correctly predicted training instance should be taken into consideration during the model training. 
In contrast, 
the context word in an incorrectly predicted training instance ought to be ignored.
To this end,
we iteratively conduct sentiment predictions on all training instances.
Particularly,
at each iteration,
we extract the context word with the maximum attention weight from each training instance to form attention supervision information,
which can be used to guide the training of attention mechanism:
in the case of correct prediction,
we will remain this word to be considered;
otherwise,
the attention weight of this word is expected to be decreased.
Then,
we mask all extracted context words of each training instance so far
and then refollow the above process to discover more 
supervision information for attention mechanisms.
Finally,
we augment the standard training objective with a regularizer,
which enforces attention distributions of these mined context words to be consistent with their expected distributions.

Our main contributions are three-fold:
%\begin{itemize}
%\setlength{\itemsep}{-2.5pt}
%\item
(1) Through in-depth analysis,
we point out the existing drawback of the attention mechanism for ASC.
%\item
(2)
We propose a novel incremental approach to automatically extract attention supervision information for neural ASC models.
To the best of our knowledge,
our work is the first attempt to explore automatic attention supervision information mining for ASC.
%\item
(3) We apply our approach to two dominant neural ASC models:
Memory Network (\textbf{MN}) \cite{Tang:EMNLP2016,Wang:ACL2018} and Transformation Network (\textbf{TNet}) \cite{Li:ACL2018}.
Experimental results on several benchmark datasets
demonstrate the effectiveness of our approach.
%\end{itemize}

\section{Background}\label{Section_Background}
In this section,
we give brief introductions to MN and TNet,
which both achieve satisfying performance and thus are chosen as the foundations of our work.
Here we
introduce some notations to facilitate subsequent descriptions:
$x$= ($x_1,x_2,...,x_N$) is the input sentence,
$t$= ($t_1,t_2,...,t_T$) is the given target aspect,
$y$, $y_p$$\in$$\{$Positive, Negative, Neutral$\}$
denote the ground-truth and the predicted sentiment, respectively.

\begin{figure}[!t]
	\centering
	\includegraphics[width=8.0cm]{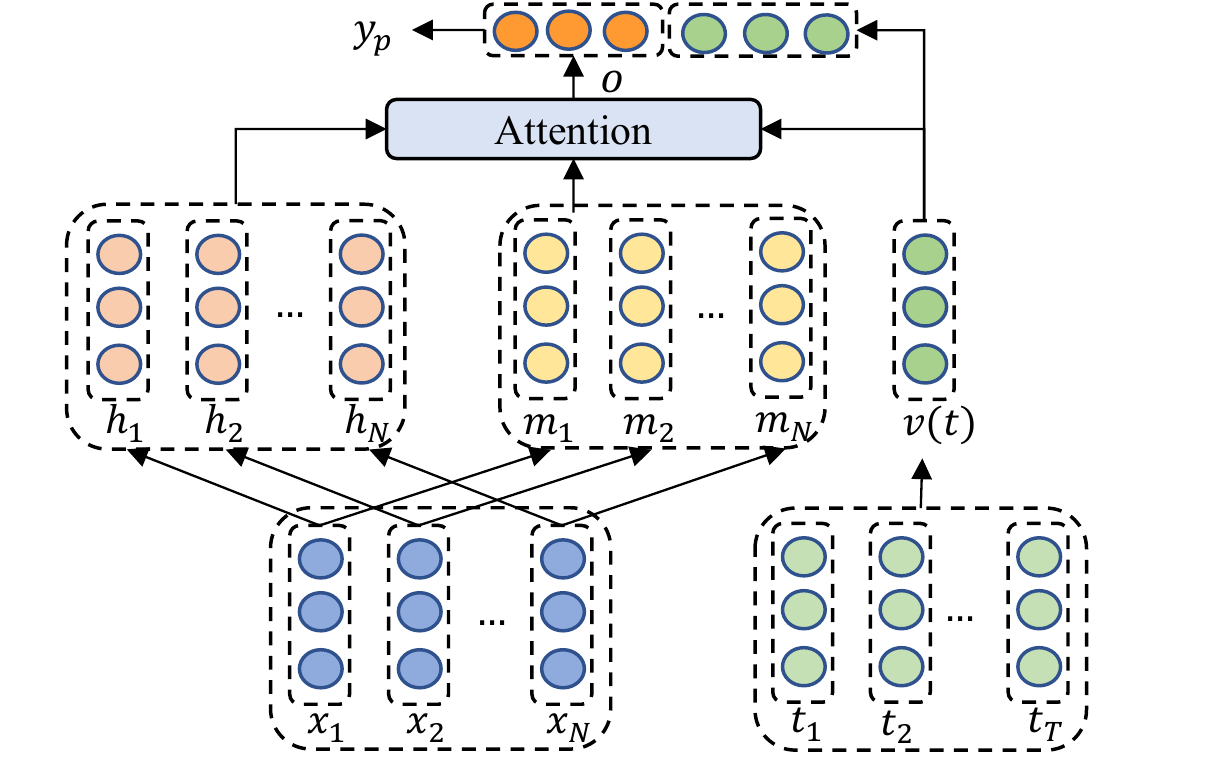}
	\caption{\label{Fig_MN}
		The framework architecture of MN.
	}
\end{figure}

\begin{figure}[!t]
	\centering
	\includegraphics[width=7.5cm]{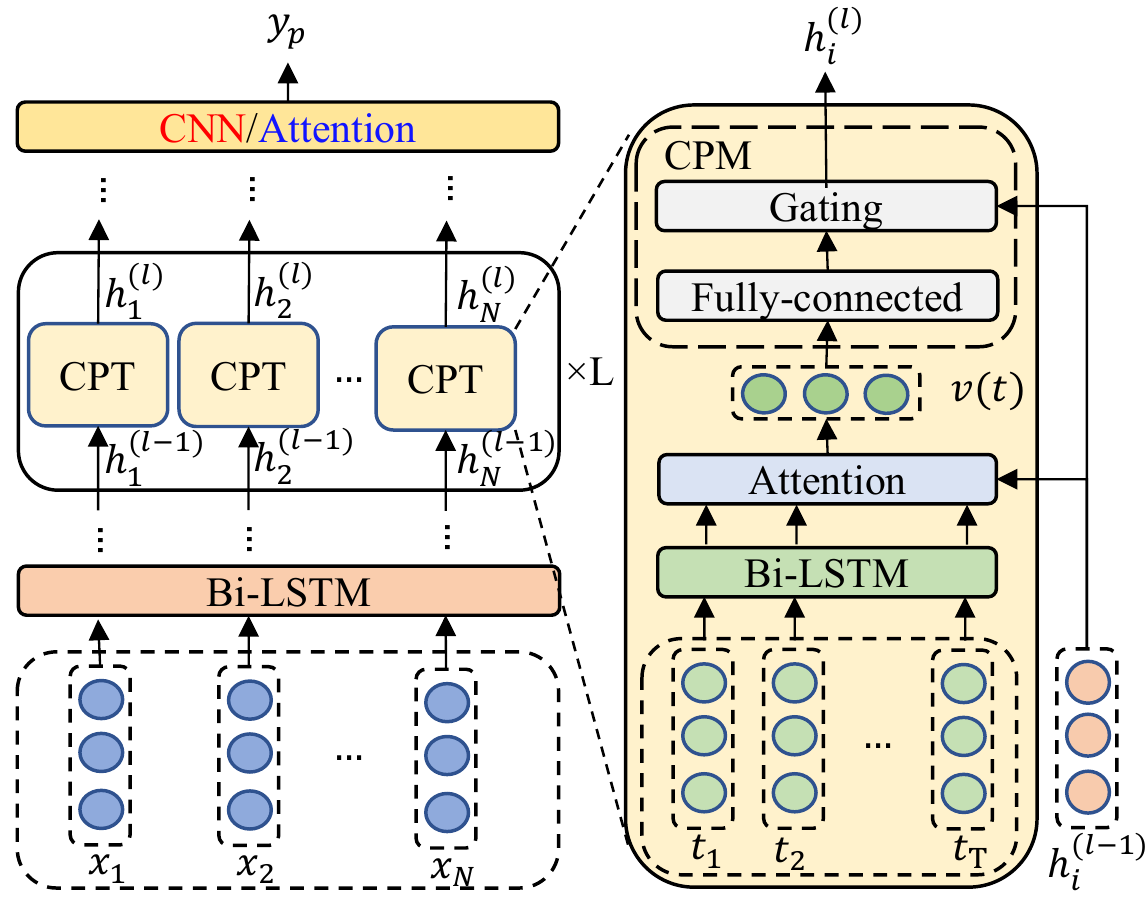}
	\caption{\label{Fig_TNet}
		The framework architecture of \textcolor[rgb]{1,0,0}{TNet}/\textcolor[rgb]{0,0,1}{TNet-ATT}.
Note that TNet-ATT is the variant of TNet replacing CNN with an attention mechanism.
	}
\end{figure}

\textbf{MN} \cite{Tang:EMNLP2016,Wang:ACL2018}.
The framework illustration of MN is given in Figure \ref{Fig_MN}.
We first introduce an aspect embedding matrix converting each target aspect word $t_j$ into a vector representation,
and then define the final vector representation $v(t)$ of $t$ as the averaged aspect embedding of its words.
Meanwhile,
another embedding matrix is used to project each context word $x_i$ to the continuous space stored in memory,
denoted by $m_i$.
Then,
an internal attention mechanism is applied to generate the aspect-related semantic representation $o$ of the sentence $x$:
$o$ =$\sum_i$\emph{softmax}$(v_t^T M m_i) h_i$,
where $M$ is an attention matrix and $h_i$ is the final semantic representation of $x_i$,
induced from a context word embedding matrix.
Finally,
we use a fully connected output layer to conduct classification based on $o$ and $v(t)$.

\textbf{TNet} \cite{Li:ACL2018}.
Figure \ref{Fig_TNet} provides the framework illustrations of TNet,
which mainly consists of three components:

(1) The bottom layer is a
Bi-LSTM that transforms the input $x$ into the contextualized word representations $h^{(0)}(x)$=($h_1^{(0)},h_2^{(0)},...,h_N^{(0)}$) (i.e. hidden states of Bi-LSTM).
(2) The middle part, as the core of the whole model,
contains $L$ layers of Context-Preserving Transformation (CPT),
where word representations are updated as $h^{(l+1)}(x)$=CPT($h^{(l)}(x)$).
The key operation of CPT layers is Target-Specific Transformation.
It contains another Bi-LSTM for generating $v(t)$ via an attention mechanism,
and then incorporates $v(t)$ into the word representations.
Besides,
CPT layers are also equipped with a Context-Preserving Mechanism (CPM) to
preserve the context information and learn more abstract word-level features.
In the end,
we obtain the word-level semantic representations $h(x)$$=$($h_1$,$h_2$...,$h_N$), 
with $h_i$$=$$h^{(L)}_i$.
(3) The topmost part is a CNN layer used to produce the aspect-related sentence representation $o$ for the sentiment classification.

In this work,
we consider another alternative to the original TNet,
which replaces its topmost CNN with an attention mechanism to produce the aspect-related sentence representation as $o$=Atten($h(x)$,\ $v(t)$).
In Section \ref{Section_Experiments},
we will investigate the performance of
the original TNet and its variant equipped with an attention mechanism,
denoted by \textbf{TNet-ATT}.

\textbf{Training Objective}.
Both of the above-mentioned models take the negative log-likelihood of the gold-truth sentiment tags as their training objectives:
\begin{align}\label{Eqa_OldTrainingObjective}
J(D;\theta) &= -\sum_{(x,t,y)\in D} J(x,t,y;\theta) \notag \\
            &=\sum_{(x,t,y)\in D} d(y) \cdot \textrm{log}d(x,t; \theta),
\end{align}where $D$ is the training corpus,
$d(y)$ is the one-hot vector of $y$,
$d(x,t;\theta)$ is the model-predicted sentiment distribution for the pair ($x$,$t$),
and $\cdot$ denotes the dot product of two vectors.

\section{Our Approach}\label{Section_OurApproach}
In this section,
we first describe the basic intuition behind our approach and then provide its details.
Finally,
we elaborate how to incorporate the mined supervision information for attention mechanisms into neural ASC models.
It is noteworthy that our method is only applied to the training optimization of neural ASC models,
without any impact on the model testing.

\subsection{Basic Intuition}\label{SubSection_BasicIntuition}

The basic intuition of our approach stems from the following fact:
in attentional ASC models,
the importance of each context word on the given aspect mainly depends on its attention weight.
Thus,
the context word with the maximum attention weight has the most important impact on the sentiment prediction of the input sentence.
Therefore,
for a training sentence,
if the prediction of ASC model is correct,
we believe that it is reasonable to continue focusing on this context word.
Conversely,
the attention weight of this context word should be decreased.

However,
as previously mentioned,
the context word with the maximum attention weight is often the one with strong sentiment polarity.
It usually occurs frequently in the training corpus
and thus tends to be overly considered during model training.
This simultaneously leads to the insufficient learning of other context words,
especially low-frequency ones with sentiment polarities.
To address this problem,
one intuitive and feasible method is to first shield the influence of this most important context word
before reinvestigating effects of remaining context words of the training instance.
In that case,
other low-frequency context words with sentiment polarities can be discovered according to their attention weights.

\subsection{Details of Our Approach}\label{SubSection_ApproachDetails}
Based on the above analysis,
we propose a novel incremental approach to automatically mine influential context words from training instances,
which can be then exploited as attention supervision information for neural ASC models.

{
\renewcommand\baselinestretch{1.15}
\begin{algorithm}[t]
\renewcommand{\algorithmicrequire}{\textbf{Input:}}\footnotesize
\renewcommand\algorithmicensure {\textbf{Return:} }\footnotesize
\caption{: Neural ASC Model Training with Automatically Mined Attention Supervision Information.}
\label{Alg_OurApproach}
\begin{algorithmic}[1]
\REQUIRE
$\textit{D}$: the initial training corpus;\\
$\theta^{init}$: the initial model parameters;\\
$\epsilon_{\alpha}$: the entropy threshold of attention weight distribution;\\
$\textit{K}$: the maximum number of training iterations;
\STATE $\theta^{(0)}$ $\leftarrow$ \textit{\textbf{Train}}($\textit{D}$;\ $\theta^{init}$)
\STATE \textbf{for} ($x$,\ $t$,\ $y$) $\in$ $\textit{D}$ \textbf{do}
\STATE \ \ \ \ \ \ $s_a(x)$ $\leftarrow$ $\emptyset$
\STATE \ \ \ \ \ \ $s_m(x)$ $\leftarrow$ $\emptyset$
\STATE \textbf{end for}
\STATE \textbf{for} $k=1,2...,\textit{K}$ \textbf{do}
\STATE \ \ \ \ \ \ $\textit{D}^{(k)}$ $\leftarrow$ $\emptyset$
\STATE \ \ \ \ \ \ \textbf{for} ($x$,\ $t$,\ $y$) $\in$ $\textit{D}$ \textbf{do}
\STATE \ \ \ \ \ \ \ \ \ \ \ \ $v(t)$ $\leftarrow$ $\textit{\textbf{GenAspectRep}}$($t$,\ $\theta^{(k-1)}$)
\STATE \ \ \ \ \ \ \ \ \ \ \ \ $x'$ $\leftarrow$ $\textit{\textbf{MaskWord}}$($x$,\ $s_a(x)$,\ $s_m(x)$)
\STATE \ \ \ \ \ \ \ \ \ \ \ \ $h(x')$  $\leftarrow$ $\textit{\textbf{GenWordRep}}$($x'$,\ $v(t)$,\ $\theta^{(k-1)}$)
\STATE \ \ \ \ \ \ \ \ \ \ \ \ $y_p$, $\alpha(x')$ $\leftarrow$ \textit{\textbf{SentiPred}}($h(x')$,\ $v(t)$,\ $\theta^{(k-1)}$)
\STATE \ \ \ \ \ \ \ \ \ \ \ \ $E(\alpha(x'))$ $\leftarrow$ \textit{\textbf{CalcEntropy}}($\alpha(x')$)
\STATE \ \ \ \ \ \ \ \ \ \ \ \ \textbf{if} $E(\alpha(x'))$ $<$ $\epsilon_{\alpha}$ \textbf{then}
\STATE \ \ \ \ \ \ \ \ \ \ \ \ \ \ \ \ \ \ \ $m$ $\leftarrow$ $argmax_{1\leq i \leq N}$ $\alpha(x'_i)$
\STATE \ \ \ \ \ \ \ \ \ \ \ \ \ \ \ \ \ \ \ \textbf{if} $y_p$ $==$ $y$ \textbf{then}
\STATE \ \ \ \ \ \ \ \ \ \ \ \ \ \ \ \ \ \ \ \ \ \ \ $s_a(x)$ $\leftarrow$ $s_a(x)$ $\cup$ $\{x'_m\}$
\STATE \ \ \ \ \ \ \ \ \ \ \ \ \ \ \ \ \ \ \ \textbf{else}
\STATE \ \ \ \ \ \ \ \ \ \ \ \ \ \ \ \ \ \ \ \ \ \ \ $s_m(x)$ $\leftarrow$ $s_m(x)$ $\cup$ $\{x'_m\}$
\STATE \ \ \ \ \ \ \ \ \ \ \ \ \ \ \ \ \ \ \ \textbf{end if}
\STATE \ \ \ \ \ \ \ \ \ \ \ \ \textbf{end if}
\STATE \ \ \ \ \ \ \ \ \ \ \ \ $\textit{D}^{(k)}$ $\leftarrow$ $\textit{D}^{(k)}$ $\cup$ ($x'$,\ $t$,\ $y$)
\STATE \ \ \ \ \ \ \textbf{end for}
\STATE \ \ \ \ \ \ $\theta^{(k)}$ $\leftarrow$ \textit{\textbf{Train}}($\textit{D}^{(k)}$;\ $\theta^{(k-1)}$)
\STATE \textbf{end for}
\STATE $\textit{D}_s$ $\leftarrow$ $\emptyset$
\STATE \textbf{for} ($x$,\ $t$,\ $y$) $\in$ $\textit{D}$ \textbf{do}
\STATE \ \ \ \ \ \ \ $\textit{D}_s$ $\leftarrow$ $\textit{D}_s$ $\cup$ ($x$,\ $t$,\ $y$,\ $s_a(x)$,\ $s_m(x)$)
\STATE \textbf{end for}
\STATE $\theta$ $\leftarrow$ \textit{\textbf{Train}}($\textit{D}_s$)
\ENSURE $\theta$;
\end{algorithmic}
\end{algorithm}
\par}

As shown in Algorithm \ref{Alg_OurApproach},
we first use the initial training corpus $D$ to conduct model training,
and then obtain the initial model parameters $\theta^{(0)}$ (\textbf{Line 1}).
Then,
we continue training the model for $K$ iterations,
where influential context words of all training instances can be iteratively extracted (\textbf{Lines 6-25}).
During this process,
for each training instance ($x,t,y$),
we introduce two word sets initialized as $\emptyset$ (\textbf{Lines 2-5}) to record its extracted context words:
(1) $s_a(x)$ consists of context words with \textbf{a}ctive effects on the sentiment prediction of $x$.
Each word of $s_a(x)$ will be encouraged to remain considered in the refined model training,
and
(2) $s_m(x)$ contains context words with \textbf{m}isleading effects,
whose attention weights are expected to be decreased.
Specifically,
at the $k$-th training iteration,
we adopt the following steps to deal with ($x,t,y$):

\begin{table*}[t]
\centering
\small
\begin{tabularx}{16cm}{|c|l|p{37pt}<{\centering}|p{38pt}<{\centering}|p{28pt}<{\centering}|}
\hline
{\bf Iter} & \multicolumn{1}{|c|}{\bf Sentence} & {\bf Ans./Pred.} & { \boldmath$E(\alpha(x'))$ \unboldmath} & \boldmath {$x'_m$} \unboldmath\\
\hline
\hline
{\bf 1} & \Appb{The}{0.06}\Appb{\bf [place]}{0.08}\Appb{is}{0.06}\Appb{small}{0.27}\Appb{and}{0.12}\Appb{crowded}{0.09}\Appb{but}{0.05}\Appb{the}{0.03}\Appb{service}{0.06}\Appb{is}{0.04}\Appb{quick}{0.13}\Appb{.}{0.01}& {\bf Neg / Neg} & {\bf 2.38} & {\bf \emph{small}}\\
\hline
{\bf 2} & \Appb{The}{0.03}\Appb{\bf [place]}{0.08}\Appb{is}{0.05}\Appb{$\langle mask\rangle$}{0.00}\Appb{and}{0.16}\Appb{crowded}{0.25}\Appb{but}{0.10}\Appb{the}{0.03}\Appb{service}{0.06}\Appb{is}{0.04}\Appb{quick}{0.17}\Appb{.}{0.01}& {\bf Neg / Neg} & {\bf 2.59} & {\bf \emph{crowded}}\\
\hline
{\bf 3} & \Appb{The}{0.01}\Appb{\bf [place]}{0.03}\Appb{is}{0.05}\Appb{$\langle mask\rangle$}{0.00}\Appb{and}{0.10}\Appb{$\langle mask\rangle$}{0.00}\Appb{but}{0.10}\Appb{the}{0.09}\Appb{service}{0.16}\Appb{is}{0.14}\Appb{quick}{0.25}\Appb{.}{0.05}& {\bf Neg / Pos} & {\bf 2.66} & {\bf \emph{quick}}\\
\hline
{\bf 4} & \Appb{The}{0.10}\Appb{\bf [place]}{0.15}\Appb{is}{0.10}\Appb{$\langle mask\rangle$}{0.00}\Appb{and}{0.11}\Appb{$\langle mask\rangle$}{0.00}\Appb{but}{0.11}\Appb{the}{0.10}\Appb{service}{0.15}\Appb{is}{0.10}\Appb{$\langle mask\rangle$}{0.00}\Appb{.}{0.05}& {\bf Neg / Neg} & {\bf 3.07} & {\bf \emph{---}}\\
\hline
\end{tabularx}
\caption{\label{Table_Example2}
The example of mining influential context words from the first training sentence in Table \ref{Table_Example1}.
$E(\alpha(x'))$ denotes the entropy of the attention weight distribution $\alpha(x')$,
$\epsilon_{\alpha}$ is entropy threshold set as $3.0$,
and $x'_m$ indicates the context word with the maximum attention weight.
Note that all extracted words are replaced with ``$\langle mask\rangle$'' and their background colors are labeled as white.
}
\end{table*}

In {\textbf{Step 1}},
we first apply the model parameters $\theta^{(k-1)}$ of the previous iteration to generate
the aspect representation $v(t)$ (\textbf{Line 9}).
Importantly,
according to $s_a(x)$ and $s_m(x)$,
we then mask all previously extracted context words of $x$ to create a new sentence $x'$,
where each masked word is replaced with a special token ``$\langle mask\rangle$'' (\textbf{Line 10}).
In this way,
the effects of these context words will be shielded
during the sentiment prediction of $x'$,
and thus other context words can be potentially extracted from $x'$.
Finally,
we generate the word representations $h(x')$$=$$\{h(x'_i)\}^N_{i=1}$ (\textbf{Line 11}).

In \textbf{Step 2},
on the basis of $v(t)$ and $h(x')$,
we leverage $\theta^{(k-1)}$ to predict the sentiment of $x'$ as $y_p$ (\textbf{Line 12}),
where the word-level attention weight distribution $\alpha(x')$=$\{\alpha(x'_1),\alpha(x'_2),...,\alpha(x'_N)\}$
subjecting to $\sum^N_{i=1}\alpha{(x'_i)}=1$ is induced.

In \textbf{Step 3},
we use the entropy $E(\alpha(x'))$ to measure the variance of $\alpha(x')$ (\textbf{Line 13}),
which contributes to determine the existence of an influential context word for the sentiment prediction of $x'$,
\begin{align}\label{Eqa_Entropy}
E(\alpha(x')) = -\sum^N_{i=1} \alpha(x'_i) \log(\alpha(x'_i)).
\end{align}If $E(\alpha(x'))$ is less than a threshold $\epsilon_{\alpha}$ (\textbf{Line 14}),
we believe that there exists at least one context word with great effect on the sentiment prediction of $x'$.
Hence,
we extract the context word $x'_m$ with the maximum attention weight (\textbf{Line 15-20})
that will be exploited as attention supervision information to refine the model training.
Specifically,
we adopt two strategies to deal with $x'_m$ according to different prediction results on $x'$:
if the prediction is correct,
we wish to continue focusing on $x'_m$ and add it into $s_a(x)$ (\textbf{Lines 16-17});
otherwise,
we expect to decrease the attention weight of $x'_m$ and thus include it into $s_m(x)$ (\textbf{Lines 18-19}).

In \textbf{Step 4},
we combine $x'$, $t$ and $y$ as a triple,
and merge it with the collected ones to form a new training corpus $D^{(k)}$ (\textbf{Line 22}).
Then,
we leverage $D^{(k)}$ to continue updating model parameters for the next iteration (\textbf{Line 24}).
In doing so,
we make our model adaptive to discover more influential context words.

Through $K$ iterations of the above steps,
we manage to extract influential context words of all training instances.
Table \ref{Table_Example2} illustrates the context word mining process of the first sentence shown in Table \ref{Table_Example1}.
In this example,
we iteratively extract three context words in turn: ``\emph{small}'', ``\emph{crowded}'' and ``\emph{quick}''.
The former two words are included in $s_a(x)$,
while the last one is contained in $s_m(x)$.
Finally,
the extracted context words of each training instance will be included into $D$,
forming a final training corpus $D_s$ with
attention supervision information (\textbf{Lines 26-29}),
which will be used to carry out the last model training (\textbf{Line 30}).
The details will be provided in the next subsection.

\subsection{Model Training with Attention Supervision Information} \label{SubSection_ModelTraining}

To exploit the above extracted context words to refine the training of attention mechanisms for ASC models,
we propose a soft attention regularizer $\triangle(\alpha(s_a(x)\cup s_m(x)), \hat{\alpha}(s_a(x)\cup s_m(x)); \theta)$ to jointly minimize the standard training objective,
where $\alpha(*)$ and $\hat{\alpha}(*)$ denotes the model-induced and expected attention weight distributions of words in $s_a(x)\cup s_m(x)$, respectively.
More specifically,
$\triangle(\alpha(*),\hat{\alpha}(*);\theta)$ is an \emph{Euclidean Distance} style loss that penalizes the disagreement between $\alpha(*)$ and $\hat{\alpha}(*)$.

As previously analyzed,
we expect to equally continue focusing on the context words of $s_a(x)$ during the final model training.
To this end,
we set their expected attention weights to the same value $\frac{1}{|s_a(x)|}$.
By doing so,
the weights of words extracted first will be reduced, and those of words extracted later will be increased,
avoiding the over-fitting of high-frequency context words with sentiment polarities and the under-fitting of low-frequency ones.
On the other hand,
for the words in $s_m(x)$ with misleading effects on the sentiment prediction of $x$,
we want to reduce their effects and thus directly set their expected weights as 0.
Back to the sentence shown in Table \ref{Table_Example2},
both ``\emph{small}'' and ``\emph{crowded}''$\in$$s_a(x)$ are assigned the same expected weight 0.5,
and the expected weight of ``\emph{quick}''$\in$$s_m(x)$ is 0.

Finally,
our objective function on the training corpus $D_s$ with attention supervision information becomes
\begin{align}\label{Eqa_NewTrainingObjective}
&J_s(D_s;\theta) = -\sum_{(x,t,y)\in D_s} \{J(x,t,y;\theta)+ \\
&\gamma\triangle(\alpha(s_a(x)\cup s_m(x)),\hat{\alpha}(s_a(x)\cup s_m(x));\theta)\}, \notag
\end{align}where $J(x,t,y;\theta)$ is the conventional training objective defined in Equation \ref{Eqa_OldTrainingObjective},
and $\gamma$$>$$0$ is a hyper-parameter that balances the preference between the conventional loss function and the regularization term.
In addition to the utilization of attention supervision information,
our method has a further advantage:
it is easier to address the vanishing gradient problem
by adding such information into the intermediate layers of the entire network \cite{Szegedy:CVPR2015}, 
because the supervision of $\hat{\alpha}(*)$ is closer to $\alpha(*)$ than $y$.

\section{Experiments}\label{Section_Experiments}
\begin{table}[t]
%\small
\centering
\begin{tabular}{c|c|c|c|c}
\hline
\textbf{Domain} & \textbf{Dataset} & \textbf{\#Pos} & \textbf{\#Neg} & \textbf{\#Neu}\\
\hline
\hline
\multirow{2}*{LAPTOP}      & Train   & 980  & 858 & 454 \\
                           & Test    & 340  & 128 & 171 \\
\hline
\multirow{2}*{REST}        & Train   & 2159 & 800 & 632 \\
		                   & Test    & 730  & 195 & 196 \\
\hline
\multirow{2}*{TWITTER}     & Train   & 1567 & 1563& 3127\\
		                   & Test    & 174  & 174 & 346 \\
\hline
\end{tabular}
\caption{\label{Table_Dataset}
Datasets in our experiments.
\textbf{\#Pos}, \textbf{\#Neg} and \textbf{\#Neu} denotes the number of instances with Positive, Negative and Neutral sentiment, respectively.
}
\end{table}

\textbf{Datasets}.
We applied the proposed approach into
MN \cite{Tang:EMNLP2016,Wang:ACL2018} and TNet-ATT \cite{Li:ACL2018} (see Section \ref{Section_Background}),
and conducted experiments on three benchmark datasets:
LAPTOP, REST \cite{Pontiki:SemEval2014} and TWITTER \cite{Dong:ACL2014}.
In our datasets,
the target aspect of each sentence has been provided.
Besides,
we removed a few instances with conflict sentiment labels as implemented in \cite{Chen:EMNLP2017}.
The statistics of the final datasets are listed in Table \ref{Table_Dataset}.

\textbf{Contrast Models}.
We referred to our two enhanced ASC models as \textbf{MN(+AS)} and \textbf{TNet-ATT(+AS)},
and compared them with
\textbf{MN}, \textbf{TNet}, and \textbf{TNet-ATT}.
Note our models require additional $K$$+$$1$-iteration training,
therefore,
we also
compared them with
the above models with additional $K$+1-iteration training,
which are denoted as \textbf{MN(+KT)}, \textbf{TNet(+KT)} and \textbf{TNet-ATT(+KT)}.
Moreover,
to investigate effects of different kinds of attention supervision information,
we also listed the performance of \textbf{MN(+AS$_a$)} and \textbf{MN(+AS$_m$)},
which only leverage context words of $s_a(x)$ and $s_m(x)$, respectively,
and the same for \textbf{TNet-ATT(+AS$_a$)} and \textbf{TNet-ATT(+AS$_m$)}.

\begin{figure}[!t]
\centering
\includegraphics[scale=0.55]{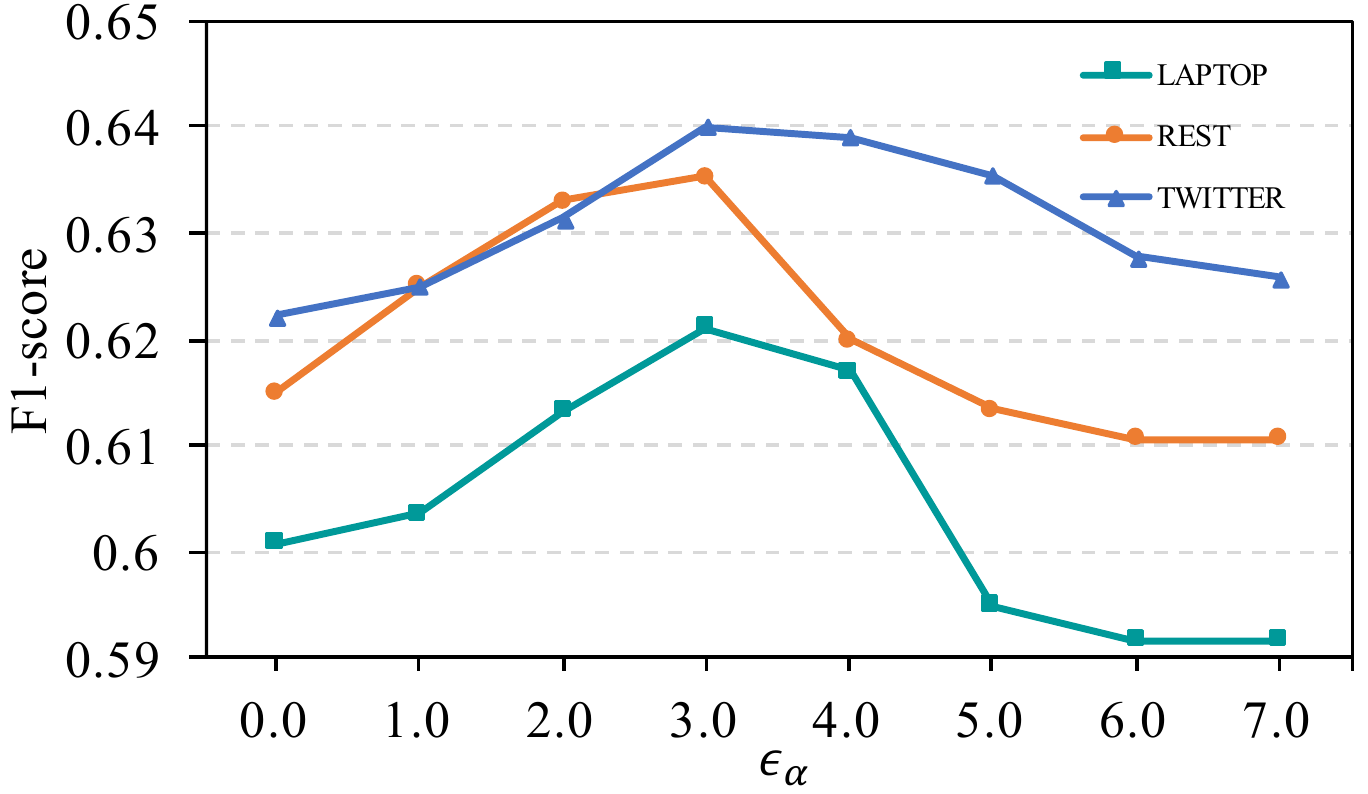}
\caption{
\label{Fig_EAEffect_MN}
Effects of $\epsilon_{\alpha}$ on the validation sets using MN(+AS).
}
\end{figure}

\begin{figure}[!t]
\centering
\includegraphics[scale=0.55]{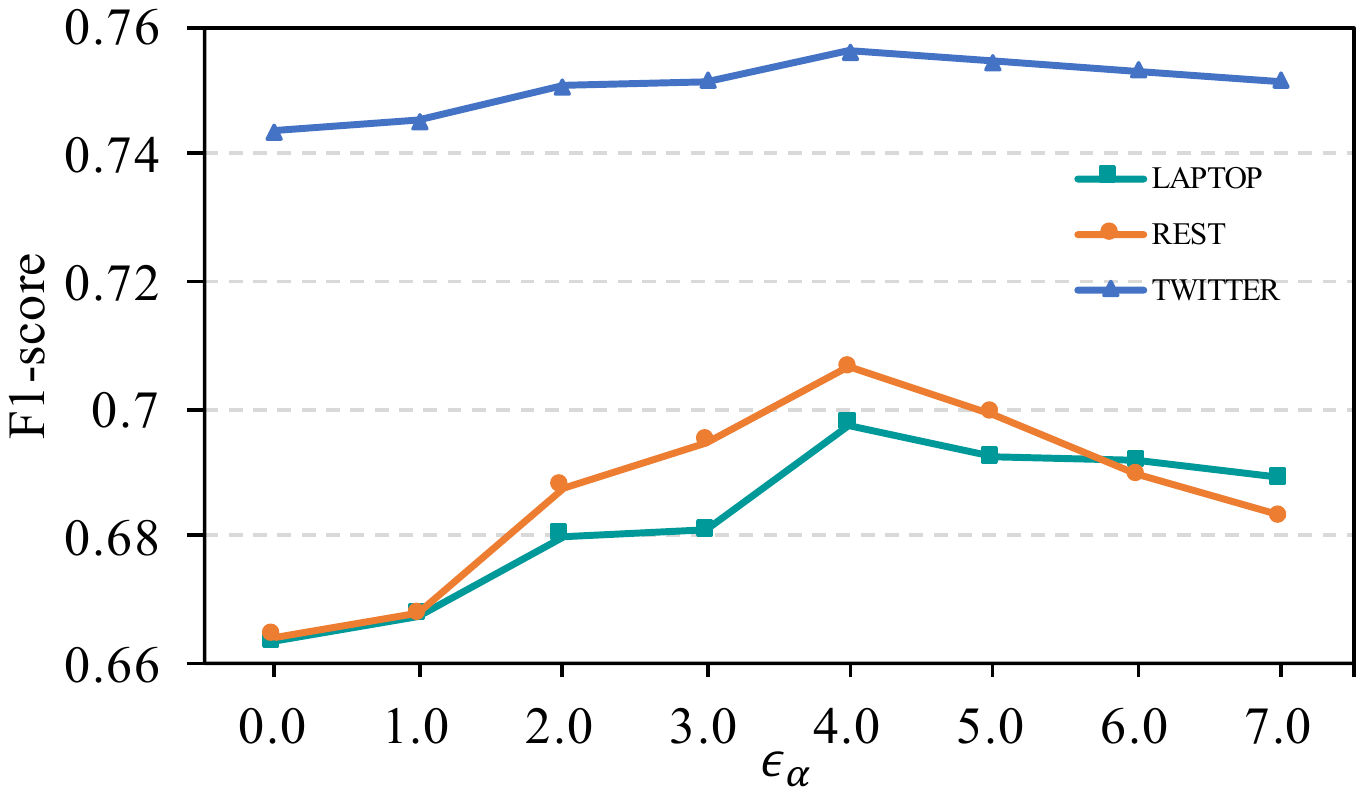}
\caption{
\label{Fig_EAEffect_TNetATT}
Effects of $\epsilon_{\alpha}$ on the validation sets using TNet-ATT(+AS).
}
\end{figure}

\begin{table*}[t]
\renewcommand\baselinestretch{1.1}
\centering
%\small
\begin{tabular}{ c|c|c|c|c|c|c }
\hline
\multirow{2}*{\textbf{Model}} & \multicolumn{2}{c|}{\textbf{LAPTOP}} & \multicolumn{2}{c|}{\textbf{REST}} & \multicolumn{2}{c}{\textbf{TWITTER}} \\
\cline{2-7}
 & Macro-F1 & Accuracy & Macro-F1 & Accuracy & Macro-F1 & Accuracy \\
\hline
\hline
MN \cite{Wang:ACL2018} & 62.89 & 68.90 & 64.34 & 75.30 & --- & --- \\
MN                     & 63.28 & 68.97 & 65.88 & 77.32 & 66.17 & 67.71 \\
MN(+KT)                & 63.31 & 68.95 & 65.86 & 77.33 & 66.18 & 67.78 \\
\hline
MN(+AS$_m$)            & 64.37 & 69.69 & 68.40 & 78.13 & 67.20 & 68.90 \\
MN(+AS$_a$)            & 64.61 & 69.95 & 68.59 & 78.23 & 67.47 & 69.17 \\
MN(+AS)                & \textbf{65.24}$^{**}$ & \textbf{70.53}$^{**}$ & \textbf{69.15}$^{**}$ & \textbf{78.75}$^{*}$ & \textbf{67.88}$^{**}$ & \textbf{69.64}$^{**}$ \\
\hline
\hline
TNet \cite{Li:ACL2018} & 71.75 & 76.54 & 71.27 & 80.69 & 73.60 & 74.97 \\
TNet                   & 71.82 & 76.12 & 71.70 & 80.35 & 76.82 & 77.60 \\
TNet(+KT)              & 71.74 & 76.44 & 71.36 & 80.59 & 76.78 & 77.54 \\
TNet-ATT               & 71.21 & 76.06 & 71.15 & 80.32 & 76.53 & 77.46 \\
TNet-ATT(+KT)          & 71.44 & 76.06 & 71.01 & 80.50 & 76.58 & 77.46 \\
\hline
TNet-ATT(+AS$_m$)      & 72.39 & 76.89 & 72.04 & 80.96 & 77.42 & 78.08 \\
TNet-ATT(+AS$_a$)      & 73.30 & 77.34 & 72.67 & 81.33 & 77.63 & 78.47 \\
TNet-ATT(+AS)          & \textbf{73.84}$^{**}$ & \textbf{77.62}$^{**}$ & \textbf{72.90}$^{**}$ & \textbf{81.53}$^{*}$ & \textbf{77.72}$^{**}$ & \textbf{78.61}$^{*}$ \\
\hline
\end{tabular}
\caption{\label{Table_OverallResults}
Experimental results on various datasets.
We directly cited the best experimental results of MN and TNet reported in \cite{Wang:ACL2018,Li:ACL2018}.
$\ast\ast$ and $\ast$ means significant at $p<$0.01 and $p<$0.05 over the baselines (MN, TNet) on each test set, respectively.
Here we conducted 1,000 bootstrap tests \cite{Koehn:EMNLP2004} to measure the significance in metric score differences.
}
\end{table*}

\textbf{Training Details}.
We used pre-trained \emph{GloVe} vectors \cite{Pennington:EMNLP2014} to initialize the word embeddings with vector dimension 300.
For out-of-vocabulary words,
we randomly sampled their embeddings from the uniform distribution [-0.25, 0.25],
as implemented in \cite{Kim:EMNLP2014}.
Besides,
we initialized the other model parameters uniformly between [-0.01, 0.01].
To alleviate overfitting,
we employed \emph{dropout} strategy \cite{Hinton:CS2012} on the input word embeddings of the LSTM and the ultimate aspect-related sentence representation.
\emph{Adam} \cite{Kingma:ICLR2015} was adopted as the optimizer with the learning rate 0.001.

When implementing our approach,
we empirically set the maximum iteration number $K$ as 5,
$\gamma$ in Equation \ref{Eqa_NewTrainingObjective} as 0.1 on LAPTOP data set, 0.5 on REST data set and 0.1 on TWITTER data set, respectively.
All hyper-parameters were tuned on 20\% randomly held-out training data.
Finally,
we used F1-Macro and accuracy as our evaluation measures.

\begin{table*}[t]
\centering
\small
\begin{tabularx}{14.5cm}{|c|l|X<{\centering}|c|}
\hline
{\bf Model} & \multicolumn{1}{|c|}{\bf Sentence} & {\bf Ans./Pred.}
\unboldmath\\
\hline
\hline
{\bf TNet-ATT} & \Appb{The}{0.23}\Appb{\bf [folding}{0.25}\Appb{\bf chair]}{0.17}\Appb{i}{0.15}\Appb{was}{0.02}\Appb{seated}{0.07}\Appb{at}{0.04}\Appb{was}{0.02}\Appb{uncomfortable}{0.02}\Appb{.}{0.09}& {\bf Neg / Neu}\\
\hline
{\bf TNet-ATT(+AS)} & \Appb{The}{0.02}\Appb{\bf [folding}{0.02}\Appb{\bf chair]}{0.02}\Appb{i}{0.02}\Appb{was}{0.04}\Appb{seated}{0.28}\Appb{at}{0.12}\Appb{was}{0.15}\Appb{uncomfortable}{0.30}\Appb{.}{0.02}& {\bf Neg / Neg}\\
\hline
\hline
{\bf TNet-ATT} & \Appb{The}{0.08}\Appb{\bf [food]}{0.12}\Appb{did}{0.12}\Appb{take}{0.08}\Appb{a}{0.08}\Appb{few}{0.08}\Appb{extra}{0.08}\Appb{minutes}{0.08}\Appb{...}{0.04}\Appb{the}{0.10}\Appb{cute}{0.24}\Appb{waiters}{0.04}\Appb{...}{0.08}& {\bf Neu / Pos}\\
\hline
{\bf TNet-ATT(+AS)} & \Appb{The}{0.02}\Appb{\bf [food]}{0.3}\Appb{did}{0.25}\Appb{take}{0.02}\Appb{a}{0.02}\Appb{few}{0.02}\Appb{extra}{0.06}\Appb{minutes}{0.06}\Appb{...}{0.02}\Appb{the}{0.02}\Appb{cute}{0.02}\Appb{waiters}{0.02}\Appb{...}{0.02}& {\bf Neu / Neu}\\
\hline
\end{tabularx}
\caption{\label{Table_Example3}
Two test cases predicted by TNet-ATT and TNet-ATT(+AS).
}
\end{table*}

\subsection{Effects of $\epsilon_{\alpha}$}
$\epsilon_{\alpha}$ is a very important hyper-parameter
that controls the iteration number of 
mining attention supervision information (see \textbf{Line 14} of Algorithm \ref{Alg_OurApproach}).
Thus,
in this group of experiments,
we varied $\epsilon_{\alpha}$ from 1.0 to 7.0 with an increment of 1 each time,
so as to investigate its effects on the performance of our models on the validation sets.

Figure \ref{Fig_EAEffect_MN} and \ref{Fig_EAEffect_TNetATT} show the experimental results of different models.
Specifically,
MN(+AS) with $\epsilon_{\alpha}$=$3.0$ achieves the best performance,
meanwhile,
the optimal performance of TNet-ATT(+AS) is obtained when $\epsilon_{\alpha}$=$ 4.0$.
We observe
the increase of $\epsilon_{\alpha}$ does not lead to further improvements,
which may be due to more noisy extracted context words.
Because of these results,
we set $\epsilon_{\alpha}$ for MN(+AS) and TNet-ATT(+AS) as 3.0 and 4.0 in the following experiments,
respectively.

\subsection{Overall Results}
Table \ref{Table_OverallResults} provides all the experimental results.
To enhance the persuasiveness of our experimental results,
we also displayed the previously reported scores of MN \cite{Wang:ACL2018} and TNet \cite{Li:ACL2018} on the same data set.
According to the experimental results,
we can come to the following conclusions:

\textbf{First},
both of our reimplemented MN and TNet are comparable to their original models reported in \cite{Wang:ACL2018,Li:ACL2018}.
These results show that our reimplemented baselines are competitive.
When we replace the CNN of TNet with an attention mechanism,
TNet-ATT is slightly inferior to TNet.
Moreover,
when we perform additional $K$+1-iteration of training on these models,
their performance has not changed significantly,
suggesting simply increasing training time is unable to enhance the performance of the neural ASC models.

\textbf{Second},
when we apply the proposed approach into both MN and TNet-ATT,
the context words in $s_a(x)$ are more effective than those in $s_m(x)$.
This is because the proportion of correctly predicted training instances is larger than that of incorrectly ones.
Besides,
the performance gap between MN(+AS$_a$) and MN(+AS$_m$) is larger than that between two variants of TNet-ATT.
One underlying reason is that the performance of TNet-ATT is better than MN,
which enables TNet-ATT to produce more correctly predicted training instances.
This in turn brings more attention supervision to TNet-ATT than MN.

\textbf{Finally},
when we use both kinds of attention supervision information,
no matter for which metric,
MN(+AS) remarkably outperforms MN on all test sets.
Although our TNet-ATT is slightly inferior to TNet,
TNet-ATT(+AS) still significantly surpasses both TNet and TNet-ATT.
These results strongly demonstrate the effectiveness and generality of our approach.

\subsection{Case Study}

In order to know how our method improves neural ASC models,
we deeply analyze attention results of TNet-ATT and TNet-ATT(+AS).
It has been found that our proposed approach can solve the above-mentioned two issues well.

Table \ref{Table_Example3} provides two test cases.
TNet-ATT incorrectly predicts the sentiment of the first test sentence as neutral.
This is because the context word ``\emph{uncomfortable}'' only appears in two training instances with negative polarities,
which distracts attention from it.
When using our approach,
the average attention weight of ``\emph{uncomfortable}'' is increased to 2.6 times than that of baseline in these two instances.
Thus,
TNet-ATT(+AS) is capable of assigning a greater attention weight (0.0056$\rightarrow$0.2940) to this context word,
leading to the correct prediction of the first test sentence.
For the second test sentence,
since the context word ``\emph{cute}'' occurs in training instances mostly with positive polarity,
TNet-ATT directly focuses on this word and then incorrectly predicts the sentence sentiment as positive.
Adopting our method,
attention weights of ``\emph{cute}'' in training instances with neural or negative polarity are significantly decreased.
Specifically, in these instances,
the average weight of ``\emph{cute}'' is reduced to 0.07 times of the original.
Hence,
TNet-ATT(+AS) assigns a smaller weight (0.1090$\rightarrow$0.0062) to ``\emph{cute}'' and achieves the correct sentiment prediction.

\section{Related Work}
Recently,
neural models have been shown to be successful on ASC.
For example,
due to its multiple advantages,
such as being simpler and faster,
MNs with attention mechanisms \cite{Tang:EMNLP2016,Wang:ACL2018} have been widely used.
Another prevailing neural model is LSTM that also involves an attention mechanism to explicitly capture the importance of each context word \cite{Wang:EMNLP2016}.
Overall,
attention mechanisms play crucial roles in all these models.

Following this trend,
researchers have resorted to more sophisticated attention mechanisms to refine neural ASC models.
Chen et al., \shortcite{Chen:EMNLP2017} proposed a multiple-attention mechanism to capture sentiment features separated by a long distance,
so that it is more robust against irrelevant information.
An interactive attention network has been designed by Ma et al., \shortcite{Ma:IJCAI2017} for ASC,
where two attention networks were introduced to model the target and context interactively.
Liu et al., \shortcite{Liu:EACL2017} proposed to leverage multiple attentions for ASC:
one obtained from the left context and
the other one acquired from the right context of a given aspect.
Very recently,
transformation-based model has also been explored for ASC \cite{Li:ACL2018},
and the attention mechanism is replaced by CNN.

Different from these work,
our work is in line with the studies of introducing attention supervision to refine the attention mechanism,
which have become hot research topics in several NN-based NLP tasks, such as
event detection \cite{Liu:ACL2017},
machine translation \cite{Liu:COLING2016},
and police killing detection \cite{Nguyen:COLING2018}.
However,
such supervised attention acquisition is labor-intense.
Therefore,
we mainly commits to automatic mining supervision information for attention mechanisms of neural ASC models.
Theoretically,
our approach is orthogonal to these models,
and we leave the adaptation of our approach into these models as future work.

Our work is inspired by two recent models:
one is \cite{Wei:CVPR2017} proposed to progressively mine discriminative object regions using classification networks to address the weakly-supervised semantic segmentation problems,
and the other one is \cite{Xu:CONLL2018}
where a dropout method integrating with global information is presented to encourage the model to mine inapparent features or patterns for text classification.
To the best of our knowledge,
our work is the first one to explore automatic mining of attention supervision information for ASC.

\section{Conclusion and Future Work}
In this paper,
we have explored how to automatically mine supervision information for attention mechanisms of neural ASC models.
Through in-depth analyses,
we first point out the defect of the attention mechanism for ASC:
a few frequent words with sentiment polarities are tend to be over-learned,
while those with low frequency often lack sufficient learning.
Then,
we propose a novel approach to automatically and incrementally mine attention supervision information for neural ASC models.
These mined information can be further used to refine the model training via a regularization term.
To verify the effectiveness of our approach,
we apply our approach into two
dominant neural ASC models,
where experimental results demonstrate our method significantly improves
the performance of these two models.

Our method is general for attention mechanisms.
Thus,
we plan to extend our approach to other neural NLP tasks with attention mechanisms,
such as neural document classification \cite{Yang:NAACL2016} and neural machine translation \cite{Zhang:TPAMI2018}.

\section*{Acknowledgments}
The authors were supported by National Natural Science Foundation of China (Nos. 61433015, 61672440), 
NSF Award (No. 1704337),
Beijing Advanced Innovation Center for Language Resources, 
the Fundamental Research Funds for the Central Universities (Grant No. ZK1024), 
Scientific Research Project of National Language Committee of China (Grant No. YB135-49), 
and Project 2019X0653 supported by XMU Training Program of Innovation and Enterpreneurship for Undergraduates. 
We also thank the reviewers for their insightful comments.

%The acknowledgments should go immediately before the references.  Do
%not number the acknowledgments section. Do not include this section
%when submitting your paper for review. \\

%\noindent \textbf{Preparing References:} \\
%Include your own bib file like this:
%\verb|\bibliographystyle{acl_natbib}|
%\verb|\bibliography{acl2019}|

%where \verb|acl2019| corresponds to a acl2019.bib file.
\bibliography{acl2019}
\bibliographystyle{acl_natbib}

%\section{Supplemental Material}
%\label{sec:supplemental}
%Submissions may include non-readable supplementary material used in the work and described in the paper. Any accompanying software and/or data should include licenses and documentation of research review as appropriate. Supplementary material may report preprocessing decisions, model parameters, and other details necessary for the replication of the experiments reported in the paper. Seemingly small preprocessing decisions can sometimes make a large difference in performance, so it is crucial to record such decisions to precisely characterize state-of-the-art methods.

%Nonetheless, supplementary material should be supplementary (rather than central) to the paper.
%\textbf{Submissions that misuse the supplementary material may be rejected without review.}
%Supplementary material may include explanations or details of proofs or derivations that do not fit into the paper, lists of features or feature templates, sample inputs and outputs for a system, pseudo-code or source code, and data.
%(Source code and data should be separate uploads, rather than part of the paper).

%The paper should not rely on the supplementary material:
%while the paper may refer to and cite the supplementary material and the supplementary material will be available to the reviewers,
%they will not be asked to review the supplementary material.

\end{document}